\documentclass[
]{ceurart}

\usepackage{listings}
\usepackage{glossaries}
\usepackage{algorithm}
\usepackage{algpseudocode}
\usepackage{wrapfig}
\usepackage{amsthm}
\usepackage{enumitem}

\theoremstyle{definition}
\newtheorem{definition}{Definition}[section]

\input{main.gls}

\begin{document}

\copyrightyear{2023}
\copyrightclause{Copyright for this paper by its authors.
  Use permitted under Creative Commons License Attribution 4.0 International (CC BY 4.0).}

\conference{SemIIM'23: 2nd International Workshop on Semantic Industrial Information Modelling, 7th November 2023, Athens, Greece, co-located with 22nd International Semantic Web Conference {(ISWC} 2023)}

\title{Semantic Association Rule Learning from Time Series Data and Knowledge Graphs}

\author[1]{Erkan Karabulut}[
orcid=0000-0003-2710-7951,
email=e.karabulut@uva.nl,
]

\author[1]{Victoria Degeler}[
orcid=0000-0001-7054-3770,
email=v.o.degeler@uva.nl,
]

\author[1]{Paul Groth}[
orcid=0000-0003-0183-6910,
email=p.t.groth@uva.nl,
]

\address[1]{University of Amsterdam, Science Park 904, Amsterdam, 1098 XH, North Holland, The Netherlands}

\cortext[1]{Corresponding author: Erkan Karabulut}

\begin{abstract}
  Digital Twins (DT) are a promising concept in cyber-physical systems research due to their advanced features including monitoring and automated reasoning. Semantic technologies such as Knowledge Graphs (KG) are recently being utilized in DTs especially for information modelling. Building on this move, this paper proposes a pipeline for semantic association rule learning in DTs using KGs and time series data. In addition to this initial pipeline, we also propose new semantic association rule criterion. The approach is evaluated on an industrial water network scenario. Initial evaluation shows that the proposed approach is able to learn a high number of association rules with semantic information which are more generalizable. The paper aims to set a foundation for further work on using semantic association rule learning especially in the context of industrial applications. 
\end{abstract}


\maketitle

\section{Introduction} 

Learning rules and patterns from data is one of the core sub-fields in data analysis and machine learning. \gls{ARM} is one specific task in rule learning where the goal is to learn association rules between variables which describe how two or more variables are associated to each other. In an industrial \gls{IoT} setting, \gls{ARM} methods are used to learn rules from time series sensor data~\cite{sunhare2022internet, degeler2014itemset}.

Association rules are in the form of $X \rightarrow Y$, which means if X holds, then Y also holds where X and Y can be a single or multiple variables with a truth value. In order to apply \gls{ARM} methods to numerical data, many approaches have been proposed including discretization, optimization and statistical approaches, under the name Numerical \gls{ARM} or Quantitative \gls{ARM}~\cite{kaushik2023numerical}. As an example, for time series data produced by IoT devices, a simple association rule based on a statistical approach can be in the form of $\text{mean}(m_{t}^X) \rightarrow \text{mean}(n_{t}^Y)$, which is interpreted as \textit{``in a time frame t, if mean measurement from sensor X is m, then the mean value of the measurements of sensor Y must be n''}. Discretization methods refer to discretizing numerical data before running an ARM algorithm. Optimization methods refer to evolutionary algorithms where rule quality criteria are generally used as fitness functions to learn rules in a desired format~\cite{kaushik2023numerical}. 

We hypothesize that for numerical data produced in industrial IoT networks, incorporating semantics of the system components in rule learning can be beneficial including discovering previously unknown relations, e.g. higher number of rules, and helping to generalize association rules of the above form. This hypothesis is tested in a specific type of IoT scenario, a \gls{DT}. \gls{DT}s have many different proposed definitions over the past two decades~\cite{d2022cognitive}. The main goal is to create a precise representation (\textit{`twin'}) of a physical system, often referred as \gls{PT}, in a digital environment and to maintain a bi-directional communication in between them. Recently, semantic technologies such as ontologies and \gls{KG}s, started to be used in DTs, for system/data modeling, establishing semantic interoperability, extracting semantic relations and/or facilitating reasoning processes~\cite{karabulut2023ontologies}. 

To the best of our knowledge, at the time of writing, there is no approach for learning rules containing semantic information as well as time series data in a DT. This study aims to fill this gap by proposing a first semantic rule learning approach utilizing KGs, based on the well-known FP-Growth~\cite{han2000mining} algorithm. Concretely, the contributions of this paper are as follows:

\begin{itemize}[noitemsep,topsep=2pt,parsep=0pt,partopsep=0pt]
    \item Describing a full pipeline of operations that consists of: i) KG construction in DTs, ii) semantic association rule learning based on the KG and time series data, and iii) making inferences based on the learned rules (Section \ref{sec:pipeline}). 
    \item A first approach (Naive SemRL) that extends FP-Growth algorithm to learn rules containing semantic information from KGs and time-series data (Section \ref{sec:naive-approach}).
    \item A semantic rule quality measure in order to evaluate rules generated by semantic rule learning algorithms (Section \ref{sec:evaluation}). 
\end{itemize}  

The proposed approach is evaluated in an industrial use case, water networks, which refer to water distribution systems that bring water to consumers, i.e. apartments, industrial sites. This is an ongoing research with many open issues and research questions emphasized in Section \ref{sec:future-work}.

\section{Semantic Rule Learning and Inference Pipeline} \label{sec:pipeline}

This section first motivates utilization of semantic association rule learning techniques in DTs, and then describes a pipeline of operations. 

\begin{figure}[tp]
    \centering
    \includegraphics[width=0.95\textwidth]{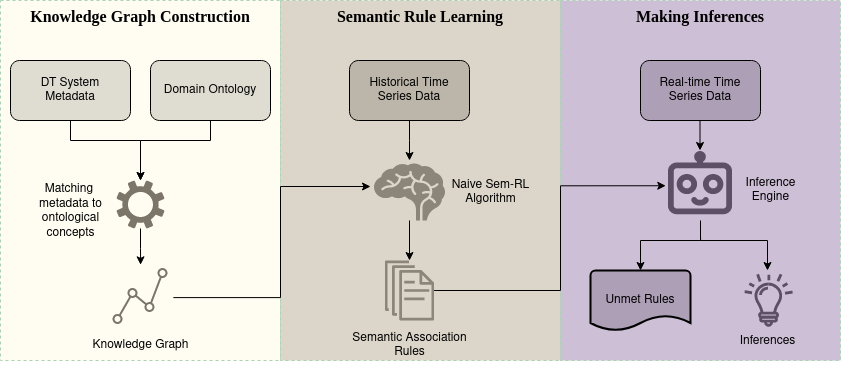}
    \caption{Semantic association rule learning and inference pipeline.}
    \label{fig:pipeline}
    \vspace{-6mm}
\end{figure}

In a best case scenario, a DT has the full knowledge of its PT. In this situation, we say that the PT is 100\% \textit{``twinned''}, or the \textit{``twinning ratio''} is 100\%. Low twinning ratio intuitively might affect the performance of any reasoning or learning algorithm running in the DT, e.g. too many missing values. A major reason that can cause low twinning ratio is to have \textit{discrepancies} in between PT and its DT. A \textit{discrepancy} refers to a state or attribute of a PT component that has incorrect or inaccurate representation in its DT. For instance, in a water network scenario, an undetected leakage in a pipe is considered a discrepancy. Instead of using separate solutions for each such issue, this study proposes a discrepancy detection method that is a generalization over any such discrepancy. The proposed approach consists of a pipeline (Figure~\ref{fig:pipeline}) of three operations: i) KG construction, ii) semantic rule learning, iii) making inferences.

\begin{figure}[bp]
    \centering
    \includegraphics[width=0.95\textwidth]{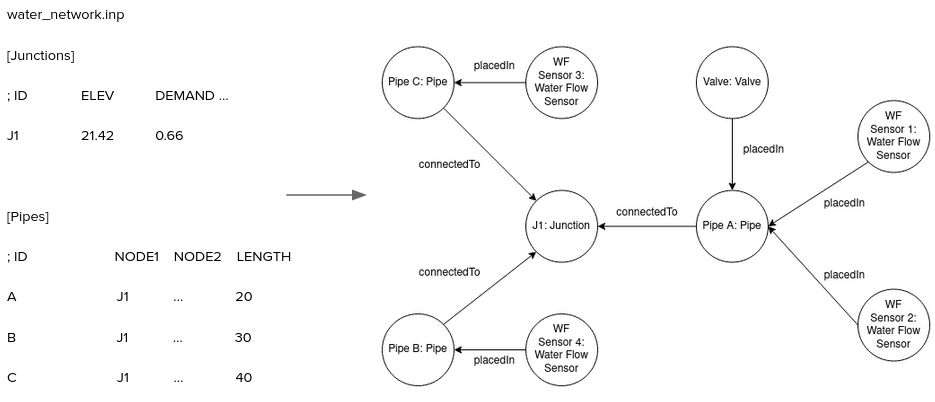}
    \caption{KG construction example for a drinking water network scenario.}
    \label{fig:kg-construction}
    \vspace{-6mm}
\end{figure}

\textbf{Knowledge Graph Construction in DTs.} KGs are already being used for information modeling in DTs~\cite{akroyd2021universal}. We hypothesize that DTs with high dependency among its sub-components, e.g. DT of a water network, can benefit from KGs, not only in information modeling but also in rule learning and making inferences. KGs for DTs can be constructed using a top-down approach from DT metadata, and a domain ontology that shows how the components are related to each other~\cite{tamavsauskaite2023defining}. Types of entities in DTs are never obscure, meaning that when a representation for a physical object is created in the digital environment, e.g. a water pipe, the type of the object is also explicitly or implicitly assigned, e.g. by putting metadata in a \textit{``water\_pipe''} table. Then, an ontology or a data schema can be used to label each entity and the relations. Figure~\ref{fig:kg-construction} shows KG construction from a partial water network metadata given in EPANET input\footnote{https://www.epa.gov/water-research/epanet} format.

\textbf{Semantic Rule Learning} refers to learning association rules with semantic information so that the learned rules are not only applicable to specific entities, but applicable to a set of entities with certain characteristics. The proposed semantic rule learning algorithm Naive SemRL, described in Section \ref{sec:naive-approach}, utilizes a KG constructed in the previous step, and historical time series data. A simple example of a rule that does not contain semantic information is \textit{`if sensor1 measures V1, then sensor2 measures V2'}. The goal of the proposed approach is to generate rules in the form of \textit{`if a sensor with type T placed in a pipe P1 with attribute1 > A1 measures V1, then the sensor with type T2 that is placed in a Junction J1 connected to P1 measures V2'}.

\textbf{Making Inferences Based on Semantic Rules.} In this phase of the pipeline, real-time time series data is analyzed based on the previously obtained semantic association rules. An inference engine gathers the rules that are not met, e.g. for a certain period of time, and makes inferences based on the unmet rules. The methodology to be used in this step remains to be among future work, while the focus of this study is on the first and second phases of the pipeline. 

\section{A Naive Semantic Rule Learning Approach - Naive SemRL} \label{sec:naive-approach}

The main intuition behind the proposed approach, Algorithm \ref{alg:naive-semrl}, is that instead of learning association rules for individual sensors, it generalizes sensor data using its metadata from the KG. The algorithm does that by extending transactions in a transaction database with semantic information extracted from the KG. As an example, rather than seeing sensor data as \textit{`\textbf{sensor X} measured value Y in a time frame T'}, it generalizes to \textit{'\textbf{a sensor} with these attributes and neighboring components measured value Y in a time frame T`}.

Naive SemRL requires a KG (\textit{knowledge\_graph}), discretized historical time series data (\textit{disc\_hist\_time\_series}, from now on `TS'), and number of neighbors (\textit{k\_neighbors}) to be analysed for semantic relations as input. TS is a set of transactions where each transaction contains discretized sensor data (items) for a certain time frame. It goes through each of the transaction in TS (lines 3-7), and first finds the topology of the items based on the k\_neighbors variable (line 4). As an example, in Figure~\ref{fig:kg-construction}, value of the topology variable for \textit{J1:Junction} would be \textit{[Pipe\_C\_ConnectedTo\_J1, Pipe\_B\_ConnectedTo\_J1, Pipe\_A\_ConnectedTo\_J1]}, assuming the value of k\_neighbors is equal to 1. A list of attributes for each of the components and links are extracted in line 5, and a new transaction is created in line 6. FP-Growth algorithm is run with the new set of transactions to discover association rules with semantic info in line 8. 

\begin{algorithm}[t]
	\caption{Naive SemRL} 
    \begin{algorithmic}[1]
    \Procedure{NaiveSemRL}{knowledge\_graph, disc\_hist\_time\_series, k\_neighbors}
        \State enriched\_transactions = []
        \For {transaction in disc\_hist\_time\_series}
            \State topology = graph.topology(transaction.sensor\_list(), k\_neighbors)
            \State attributes = graph.attributes(transaction.sensor\_list(), k\_neighbors)
            \State enriched\_transactions.append(transaction + topology + attributes)
        \EndFor
        \State \Return FP-Growth(enriched\_transactions)
    \EndProcedure
    \end{algorithmic}
    \label{alg:naive-semrl}
\end{algorithm}

\section{Preliminary Evaluation and Industrial Use Case} \label{sec:evaluation}

This section presents a semantic rule quality criteria that measures generalizability of semantic association rules, and an industrial use case where the proposed approach is applied.

\subsection{A Semantic Rule Quality Criterion}

Many quality criteria for association rules are proposed with the most fundamental ones being support and confidence~\cite{kaushik2023numerical}. However, we were unable to identify an association rule quality criterion that is specific to evaluating the semantic aspect of the learned rules. Therefore, we propose the following \textit{``semantic expressivity''} association rule quality measure:

\begin{definition}[Semantic Expressivity] \label{def:semantic-expressivity}
Let $C=\{c_1, c_2, ..., c_n\}$ be a set of classes in an ontology (or a data schema). $\forall c \in C (\text{has\_attributes}(c, \{a_1^c, a_2^c, ..., a_m^c\}))$, with $\text{has\_attributes}(x, a_m^x)$ = class x has set of $a_m^x$ attributes, and assume $\text{attributes}(x)$ = number of attributes in class x. Let $I=\{i_1, i_2, ..., i_b\}$ be a set of items. $\forall i \in I (i = (a^c \# z))$, where $a^c$ = an attribute of a class c, \# = any comparison operation, and z any value. $X \rightarrow Y$ is an implication (association rule) where $X, Y \subseteq I$. Finally, let $\text{instances}(X)$ be the different class instances in X and let $\text{attr\_count}(X, c)$ be number of items which have attributes of instance of a class c in X.

$\text{attr\_ratio}(X) =  \prod_{p}^{\text{instances}(X)} \frac{\text{attr\_count}(X, p)}{\text{attributes}(p)}$

$\text{Semantic\_Expressivity}(X \rightarrow Y) = \frac{(1 - \text{attr\_ratio}(X)) \times (1 - \text{attr\_ratio}(Y))}{(\text{instances}(X) + \text{instances}(Y)) / 2}$
    
\end{definition}

\textbf{Intuition.} Learned rules may contain different levels of semantic information. Including too much semantics in the rule makes it less generalizable, hence less \textit{`semantically expressive'}. For instance, \textit{``Junctions with 3 pipes have 1500-2000Pa water pressure''} is more general than \textit{``Junctions with 3 pipes where each of the pipes is 50-100m long, and has 2-3m diameter have 1500-2000Pa water pressure''}. The main goal of the proposed quality criterion is to understand how semantically expressive a rule is by giving each rule a score between 0 and 1.  Assume \textit{X} in $X \rightarrow Y$ is `$\{pipe\_diameter > 2\}$', with \textit{pipe} being a class in a water network ontology that can have 3 attributes: diameter, length and elevation. In this case $attr\_ratio(X) = 1/3$ since X is about one of the attributes only. Low attribute ratio leads to high semantic expressivity as the formula includes $(1 - attr\_ratio(X)) \times (1 - attr\_ratio(Y))$. Average number of instances is included in the divisor part for the purpose of incorporating topology into the formula. As an example, an association rule about a node with 3 neighbors will have a higher divisor than a node with 2 neighbors which is more general.

\subsection{Industrial Use Case}

The proposed algorithm is demonstrated on LeakDB dataset~\cite{vrachimis2018leakdb}, an artificially created realistic leakage dataset for water distribution networks. It contains metadata of 31 junctions, 1 reservoir, 34 pipes and 1,716,960 sensor measurements. Water network related classes in EPYNET Python package\footnote{https://github.com/Vitens/epynet} is used as a data schema while creating a KG. MLxtend's~\cite{raschkas_2018_mlxtend} FP-Growth implementation is used while implementing the Naive SemRL algorithm. For simplicity, the proposed approach tested using a straightforward discretization method of lowering sensor measurement precisions and calculating daily averages.

\textbf{A sample rule learned from the described dataset:} \textit{`\{(WaterPressureSensor: WPS, placed\_in, Junction: J1), (Junction: J1, connected\_to, Pipe: P1), (WaterPressureSensor: WPS, measures, 43)\} $\rightarrow$  \{(WaterConsumptionSensor: WCS, placed\_in, Junction: J2), (Junction: J2, connected\_to, Pipe: P2), (Junction: J2, connected\_to, Pipe: P3), (WaterConsumptionSensor: WCS, measures, 38)\}'}.

Interpretation of the rule: \textit{`When there is a water pressure sensor WPS placed inside a junction J1, and J1 is connected to a Pipe P1, and WPS measures 43, then a water consumption sensor WCS placed in a Junction J2 that is connected to Pipes P2 and P3 must measure 38'}. The semantic expressivity of the rule is 0.28, as it does not contain any attribute and based on 3 instances on the antecedents side, and 4 instances on the consequents side. In this experiment, Naive SemRL is run with topology info only without node attributes, as it increases runtime of the algorithm exponentially. Currently, an intuition/search-based approach is investigated that can tell when and where to include semantics in order to avoid exponential increase in the runtime. Table \ref{table:results} shows how incorporating semantics in the rule learning process increases the number of rules, together with min and max semantic expressivity values. Besides finding new rules about the same nodes extended with semantics, when run with low support thresholds, Naive SemRL can find new rules which FP-Growth can not find. Having more rules is not necessarily good as it may increase the time required for post-processing and making inferences. In order to overcome this hurdle, incorporating semantics within evolutionary or other approaches that can directly learn rules satisfying certain rule quality criteria is being investigated as part of future work.\looseness=-1 

\begin{table}[tp]
\caption{Effect of support threshold on the number of rules and semantic expressivity (SE) (confidence = 0.9)}
\begin{tabular}{|c|c c c|c|}
 \hline
         & \multicolumn{3}{|c|}{Naive SemRL} & \multicolumn{1}{|c|}{FP-Growth} \\ \hline 
        Support & Number of Rules & Max SE & Min SE & Number of Rules \\ \hline 
        0.2 & 7819 & 0.66 & 0.14 & 28 \\
        0.3 & 816 & 0.66 & 0.22 & 9\\
        0.4 & 50 & 0.66 & 0.33 & 2 \\
        0.5 & 50 & 0.66 & 0.33 & 2 \\ \hline
\end{tabular}
\label{table:results}
\vspace{-4mm}
\end{table}

\section{Conclusions and Future Work} \label{sec:future-work}

This study proposed a semantic association rule learning pipeline for Digital Twins. The pipeline consists of knowledge graph construction, semantic association rule learning from knowledge graphs and time series data, and making inferences. An initial naive approach for semantic rule learning, Naive SemRL, and a first semantic rule quality criterion is proposed. The new approach is evaluated in a water network use case and the results show that incorporating knowledge graphs allows us to learn rules with semantic information which are more generalizable.

There are many open issues and research questions yet to be answered. KG construction for DTs from system metadata and a domain ontology will be automatized. FP-Growth will be replaced by novel rule learning methods with different perspectives, e.g. statistical vs. optimization-based NARM methods. And these methods will be compared based on applicability of the proposed approach and quality criterion. Lastly, an inference mechanism that can detect and find root-causes of discrepancies from semantic rules will be developed.

\newpage

\begin{acknowledgments}
This work has received support from The Dutch Research Council (NWO), in the scope of Digital Twin for Evolutionary Changes in water networks (DiTEC) project, file number 19454. We would like to thank Vitens N.V. for providing us a historical water network sensor dataset to test our approach.
\end{acknowledgments}

\bibliography{main}

\begin{thebibliography}{10}
\expandafter\ifx\csname natexlab\endcsname\relax\def\natexlab#1{#1}\fi
\providecommand{\url}[1]{\texttt{#1}}
\providecommand{\href}[2]{#2}
\providecommand{\path}[1]{#1}
\providecommand{\DOIprefix}{doi:}
\providecommand{\ArXivprefix}{arXiv:}
\providecommand{\URLprefix}{URL: }
\providecommand{\Pubmedprefix}{pmid:}
\providecommand{\doi}[1]{\href{http://dx.doi.org/#1}{\path{#1}}}
\providecommand{\Pubmed}[1]{\href{pmid:#1}{\path{#1}}}
\providecommand{\bibinfo}[2]{#2}
\ifx\xfnm\relax \def\xfnm[#1]{\unskip,\space#1}\fi
\bibitem[{Sunhare et~al.(2022)Sunhare, Chowdhary, and
  Chattopadhyay}]{sunhare2022internet}
\bibinfo{author}{P.~Sunhare}, \bibinfo{author}{R.~R. Chowdhary},
  \bibinfo{author}{M.~K. Chattopadhyay},
\newblock \bibinfo{title}{Internet of things and data mining: An application
  oriented survey},
\newblock \bibinfo{journal}{Journal of King Saud University-Computer and
  Information Sciences} \bibinfo{volume}{34} (\bibinfo{year}{2022})
  \bibinfo{pages}{3569--3590}.
\bibitem[{Degeler et~al.(2014)Degeler, Lazovik, Leotta, and
  Mecella}]{degeler2014itemset}
\bibinfo{author}{V.~Degeler}, \bibinfo{author}{A.~Lazovik},
  \bibinfo{author}{F.~Leotta}, \bibinfo{author}{M.~Mecella},
\newblock \bibinfo{title}{Itemset-based mining of constraints for enacting
  smart environments},
\newblock in: \bibinfo{booktitle}{2014 IEEE International Conference on
  Pervasive Computing and Communication Workshops (Percom Workshops)},
  \bibinfo{year}{2014}, pp. \bibinfo{pages}{41--46}.
  \DOIprefix\doi{10.1109/PerComW.2014.6815162}.
\bibitem[{Kaushik et~al.(2023)Kaushik, Sharma, Fister~Jr, and
  Draheim}]{kaushik2023numerical}
\bibinfo{author}{M.~Kaushik}, \bibinfo{author}{R.~Sharma},
  \bibinfo{author}{I.~Fister~Jr}, \bibinfo{author}{D.~Draheim},
\newblock \bibinfo{title}{Numerical association rule mining: A systematic
  literature review},
\newblock \bibinfo{journal}{arXiv preprint arXiv:2307.00662}
  (\bibinfo{year}{2023}).
\bibitem[{D’Amico et~al.(2022)D’Amico, Erkoyuncu, Addepalli, and
  Penver}]{d2022cognitive}
\bibinfo{author}{R.~D. D’Amico}, \bibinfo{author}{J.~A. Erkoyuncu},
  \bibinfo{author}{S.~Addepalli}, \bibinfo{author}{S.~Penver},
\newblock \bibinfo{title}{Cognitive digital twin: An approach to improve the
  maintenance management},
\newblock \bibinfo{journal}{CIRP Journal of Manufacturing Science and
  Technology} \bibinfo{volume}{38} (\bibinfo{year}{2022})
  \bibinfo{pages}{613--630}.
\bibitem[{Karabulut et~al.(2023)Karabulut, Pileggi, Groth, and
  Degeler}]{karabulut2023ontologies}
\bibinfo{author}{E.~Karabulut}, \bibinfo{author}{S.~F. Pileggi},
  \bibinfo{author}{P.~Groth}, \bibinfo{author}{V.~Degeler},
  \bibinfo{title}{Ontologies in digital twins: A systematic literature review},
  \bibinfo{year}{2023}. \href{http://arxiv.org/abs/2308.15168}{{\tt
  arXiv:2308.15168}}.
\bibitem[{Han et~al.(2000)Han, Pei, and Yin}]{han2000mining}
\bibinfo{author}{J.~Han}, \bibinfo{author}{J.~Pei}, \bibinfo{author}{Y.~Yin},
\newblock \bibinfo{title}{Mining frequent patterns without candidate
  generation},
\newblock \bibinfo{journal}{ACM sigmod record} \bibinfo{volume}{29}
  (\bibinfo{year}{2000}) \bibinfo{pages}{1--12}.
\bibitem[{Akroyd et~al.(2021)Akroyd, Mosbach, Bhave, and
  Kraft}]{akroyd2021universal}
\bibinfo{author}{J.~Akroyd}, \bibinfo{author}{S.~Mosbach},
  \bibinfo{author}{A.~Bhave}, \bibinfo{author}{M.~Kraft},
\newblock \bibinfo{title}{Universal digital twin-a dynamic knowledge graph},
\newblock \bibinfo{journal}{Data-Centric Engineering} \bibinfo{volume}{2}
  (\bibinfo{year}{2021}) \bibinfo{pages}{e14}.
\bibitem[{Tama{\v{s}}auskait{\.e} and Groth(2023)}]{tamavsauskaite2023defining}
\bibinfo{author}{G.~Tama{\v{s}}auskait{\.e}}, \bibinfo{author}{P.~Groth},
\newblock \bibinfo{title}{Defining a knowledge graph development process
  through a systematic review},
\newblock \bibinfo{journal}{ACM Transactions on Software Engineering and
  Methodology} \bibinfo{volume}{32} (\bibinfo{year}{2023})
  \bibinfo{pages}{1--40}.
\bibitem[{Vrachimis et~al.(2018)Vrachimis, Kyriakou
  et~al.}]{vrachimis2018leakdb}
\bibinfo{author}{S.~G. Vrachimis}, \bibinfo{author}{M.~S. Kyriakou}, et~al.,
\newblock \bibinfo{title}{Leakdb: a benchmark dataset for leakage diagnosis in
  water distribution networks:(146)},
\newblock in: \bibinfo{booktitle}{WDSA/CCWI Joint Conference Proceedings},
  volume~\bibinfo{volume}{1}, \bibinfo{year}{2018}.
\bibitem[{Raschka(2018)}]{raschkas_2018_mlxtend}
\bibinfo{author}{S.~Raschka},
\newblock \bibinfo{title}{Mlxtend: Providing machine learning and data science
  utilities and extensions to python’s scientific computing stack},
\newblock \bibinfo{journal}{The Journal of Open Source Software}
  \bibinfo{volume}{3} (\bibinfo{year}{2018}). \URLprefix
  \url{https://joss.theoj.org/papers/10.21105/joss.00638}.
  \DOIprefix\doi{10.21105/joss.00638}.

\end{thebibliography}

\appendix

\end{document}